\documentclass[sigconf,natbib=false]{acmart}

\AtBeginDocument{%
  }

\RequirePackage[
  datamodel=acmdatamodel,
  style=acmnumeric,
  ]{biblatex}

\begin{document}

\title{FusionMind - Improving question and answering with external context fusion}

\author{Shreyas Verma}
\affiliation{%
  \institution{Georgia Institute of Technology}
  \streetaddress{Atlanta, Georgia}
  \city{Atlanta}
  \state{Georgia}
  \country{USA}
  }

\author{Manoj Parmar}
\affiliation{%
  \institution{Georgia Institute of Technology}
  \streetaddress{Atlanta, Georgia}
  \city{Atlanta}
  \state{Georgia}
  \country{USA}
  }

\author{Palash Choudhary}
\affiliation{%
  \institution{Georgia Institute of Technology}
  \streetaddress{Atlanta, Georgia}
  \city{Atlanta}
  \state{Georgia}
  \country{USA}
  }

\author{Sanchita Porwal}
\affiliation{%
  \institution{Georgia Institute of Technology}
  \streetaddress{Atlanta, Georgia}
  \city{Atlanta}
  \state{Georgia}
  \country{USA}
  }


\begin{abstract}
Answering questions using pre-trained language models (LMs) and knowledge graphs (KGs) presents challenges in identifying relevant knowledge and performing joint reasoning. We compared LMs (fine-tuned for the task) with the previously published QAGNN method for the Question-answering (QA) objective and further measured the impact of additional factual context on the QAGNN performance. The QAGNN method employs LMs to encode QA context and estimate KG node importance, and effectively update the question choice entity representations using Graph Neural Networks (GNNs). We further experimented with enhancing the QA context encoding by incorporating relevant knowledge facts for the question stem. The models are trained on the OpenbookQA dataset, which contains \textasciitilde6000 4-way multiple choice questions and is widely used as a benchmark for QA tasks.
Through our experimentation, we found that incorporating knowledge facts context led to a significant improvement in performance. In contrast, the addition of knowledge graphs to language models resulted in only a modest increase. This suggests that the integration of contextual knowledge facts may be more impactful for enhancing question-answering performance compared to solely adding knowledge graphs.
\end{abstract}



\maketitle

\section{Introduction}

The ability to reason and answer questions using observation and knowledge is a crucial aspect of human intelligence. Consequently, reasoning has become an important task in natural language understanding. There have been multiple attempts to build QA systems for structured reasoning, including by combining LMs and KGs. While LMs are effective in language understanding, they struggle with handling knowledge and answering questions with underlying structural reasoning. They also have sub-par performances when compared to LM+KG tasks that leverage the embedded knowledge in graphs to answer questions and also improve the underlying reasoning.   

We explored question-answering by implementing and experimenting with two approaches. The first approach involved fine-tuning language models for the QA task on the OpenBookQA \cite{mihaylov2018suit} dataset. The second method involved combining LMs and KGs together. Using LMs and KGs together involves two main challenges: (i) identifying relevant knowledge from large KGs and (ii) performing joint reasoning over the QA context and KG. We experimented with the methodologies from "QAGNN: Reasoning with Language Models and Knowledge Graphs for Question Answering" \cite{qa-gnn}. This publication tackles the 2 above-mentioned challenges with two key innovations: (i) relevance scoring, which uses LMs to estimate the importance of KG nodes relative to the QA context, and (ii) joint reasoning, which connects the QA context and KG in a joint graph and mutually updates their representations through graph neural networks.

Additionally, we experimented with supplemental knowledge facts by using the knowledge facts in the OpenBookQA dataset. Drawing inspiration from \cite{fusing-context}, which leverages Wiktionary to provide descriptions for graph entities, we investigated the impact of supplementary knowledge facts on the question-answering task. 

Through our experimentation, we found that incorporating knowledge facts context led to a significant improvement in performance. In contrast, the addition of knowledge graphs to language models resulted in only a modest increase. This suggests that the integration of contextual knowledge facts may be more impactful for enhancing question-answering performance compared to solely adding knowledge graphs.

The impact of our project will be significant for researchers, AI developers, and end-users in the natural language understanding domain. By successfully enhancing the existing methodologies, our project can potentially advance the state-of-the-art in question answering, contributing to more accurate and context-aware AI systems, and improving overall user experience across various applications, including virtual assistants, search engines, and educational tools. This methodology would aid in answering complex questions with underlying structural reasoning, a task that is still wanting in pre-trained small-scale LMs like BERT. A logical extension to the work would be efficiently incorporating new disjoint context (as provided in \cite{fusing-context}) to the already existing information while keeping in mind the input token size of the LMs.

\section{LITERATURE SURVEY/BASELINES}





Pre-trained LMs have demonstrated remarkable success \cite{Gupta2023InstructionTM} in many tasks such as question-answering tasks (\cite{lm1},\cite{lm2}), Synthetic Data Generation \cite{Gupta2023TarGENTD}, Sentiment Analysis \cite{Scaria2023InstructABSAIL}, Event Detection \cite{Anantheswaran2023EDM3ED}. LLMs are also used in multiple domains like clinical data \cite{Parmar2023LongBoXET} and finance \cite{Gupta2021ContextNERC}. However, while LMs have broad coverage of knowledge, they do not empirically perform well on structured reasoning 
and enabling explainable predictions, which KGs are more suited for. Existing LM + KG methods for QA tasks (\cite{kagnet},\cite{kglm2}) treat the QA context and KG as two separate modalities. They individually apply LMs to the QA context and graph neural networks (GNNs) to the KG and do not mutually update or unify their representations. This separation might limit their capability to perform structured reasoning, e.g., handling negation. 

\cite{qa-gnn}, an improvement on the above methodology, uses LMs to encode the QA context and retrieve a KG subgraph from the QA entities. The QA context is connected with the retrieved KG to form a joint graph which is then passed through a GNN module for reasoning. This simultaneously updates the representation of both the KG entities and the QA context node, bridging the gap between the two sources of information.

Drawing inspiration from \cite{fusing-context}, which leverages Wiktionary to provide descriptions for graph entities, we further investigated the impact of additional knowledge facts from the OpenBookQA dataset on the question-answering task. While this approach has been touched upon in \cite{qa-gnn}, our objective was to expand upon these experiments to understand better the potential benefits of incorporating supplementary knowledge facts.

\section{Dataset description}

\subsection{Data preparation}





There are 2 primary datasets used for our process:

\begin{itemize}
\item OpenBookQA  \cite{mihaylov2018suit}: The data was obtained from the OpenBookQA website \href{https://allenai.org/data/open-book-qa}{[Data Link]}. This dataset provides a comprehensive set of multiple-choice questions and related knowledge facts to train and evaluate our model on elementary-level science question answering tasks. We additionally incorporate the 1,326 knowledge facts provided, to enhance the QA context vector.

\item ConceptNet \cite{speer2018conceptnet}: We accessed the ConceptNet \href{https://conceptnet.io/}{[API]} to retrieve relevant concepts and relationships, enhancing our model's knowledge base with structured semantic information. The data in ConceptNet is available in a JSON-LD API.


\end{itemize}
To preprocess the data, we performed the following steps on the OpenBookQA and ConceptNet datasets:

\begin{itemize}
\item Setup ConceptNet into a format that can directly be used by us: extracting English relations from ConceptNet and merging the original 42 relation types into 17 types.
\item Convert the QA datasets into .jsonl files.
\item Identify all mentioned concepts in the questions and answers.
\item Extract working subgraphs for each QA pair.

\end{itemize}


To summarise, we used the ConceptNet data for its knowledge graph and the OpenBookQA for its multiple-choice questions and related facts to train the language model. 

\subsection{Raw Data Statistics: Explain the dataset}


A few highlights of the OpenBookQA data: (Table 1)
\begin{itemize}
\item Important features: Each question comes with four possible answer choices, and the correct answer is labeled.
\item Ground-truth labels: The correct answer choice is provided for each question.
\end{itemize}


\begin{table}
  \caption{OpenBookQA dataset statistics}
  \label{tab:openbookqa-stats}
  \begin{tabular}{ccccc}
    \toprule
    \#Train & \#Dev & \#Test & \#Choices \\
    \midrule
     4,957 & 500 & 500 & 4 \\
  \bottomrule
\end{tabular}
\end{table}
\begin{table}
\begin{tabular}{cccccc}
    \toprule
    Avg. length of Q & Max. QA Token Size & Max. QA + Fact Token Size\\
    \midrule
    \textasciitilde 11 words & 73 & 101\\
  \bottomrule
\end{tabular}
\end{table}

A few highlights of the ConceptNet data: (Table 2)
\begin{itemize}
\item Important features: Each concept and relationship has associated properties, such as weight and frequency.
\item Number of labels/classes: There are no explicit labels or classes in ConceptNet.
\item Most frequent relationship type: "HasSubevent".
\end{itemize}

\begin{table}
  \caption{Processed ConceptNet dataset statistics}
  \label{tab:conceptnet-stats}
  \begin{tabular}{ccc}
    \toprule
    \#Nodes&\#Edges&\#Relationship types\\
    \midrule
    \textasciitilde~784K & \textasciitilde~4.3M & 17\\
  \bottomrule
\end{tabular}
\end{table}





\section{EXPERIMENTAL SETTINGS}



We evaluate the performance of our model on the OpenBookQA dataset. The dataset split is mentioned in Table 1. Since this is a supervised QA (classification) task, we use accuracy as the evaluation metric to compare different approaches as we want to see what percentage of our predictions are correct. We also experiment with various hyperparameters including the number of layers (k) in the GNN module, the number of fully connected layers before the output, and the inclusion of knowledge facts. For the training task, we primarily use Google Colab and PACE COC-ICE clusters. COC-ICE has 23 CPU nodes and multiple GPU nodes with RTX 6000 and dual Tesla V100 GPUs.

\section{METHODS}

\subsection{Approach 1: LM only method}
As described in Figure 1, for our first approach, we fine-tuned LMs for QA tasks on the OpenBookQA dataset. The LM model that we use is DistilRoberta-base \cite{lm1} to provide reliable and strong language-based baselines for comparison. We chose Distilroberta-base as our choice of LM because of the faster training time and strong performance. 
Language models fine-tuned on QA datasets \cite{rajpurkar2016squad}  have been extensively used as benchmarks for QA tasks \cite{wang2022modern} \cite{zhu2021retrieving}. Hence we think that fine-tuning the LM architecture for our specific problem statement would serve as a reliable approach to compare our second approach with.
The final model was trained on Google Colab with a Tesla K80 accelerator and has the following hyperparameter setting:

\begin{itemize}
    \item Model to finetune: distilroberta-base
    \item Batch size: 16
    \item Epochs: 20
    \item Learning rate: 5e-5
    \item Optimizer: AdamW
    \item Loss: Cross-entropy
\end{itemize}

\begin{figure}[h]
  \centering
  \includegraphics[width=\linewidth]{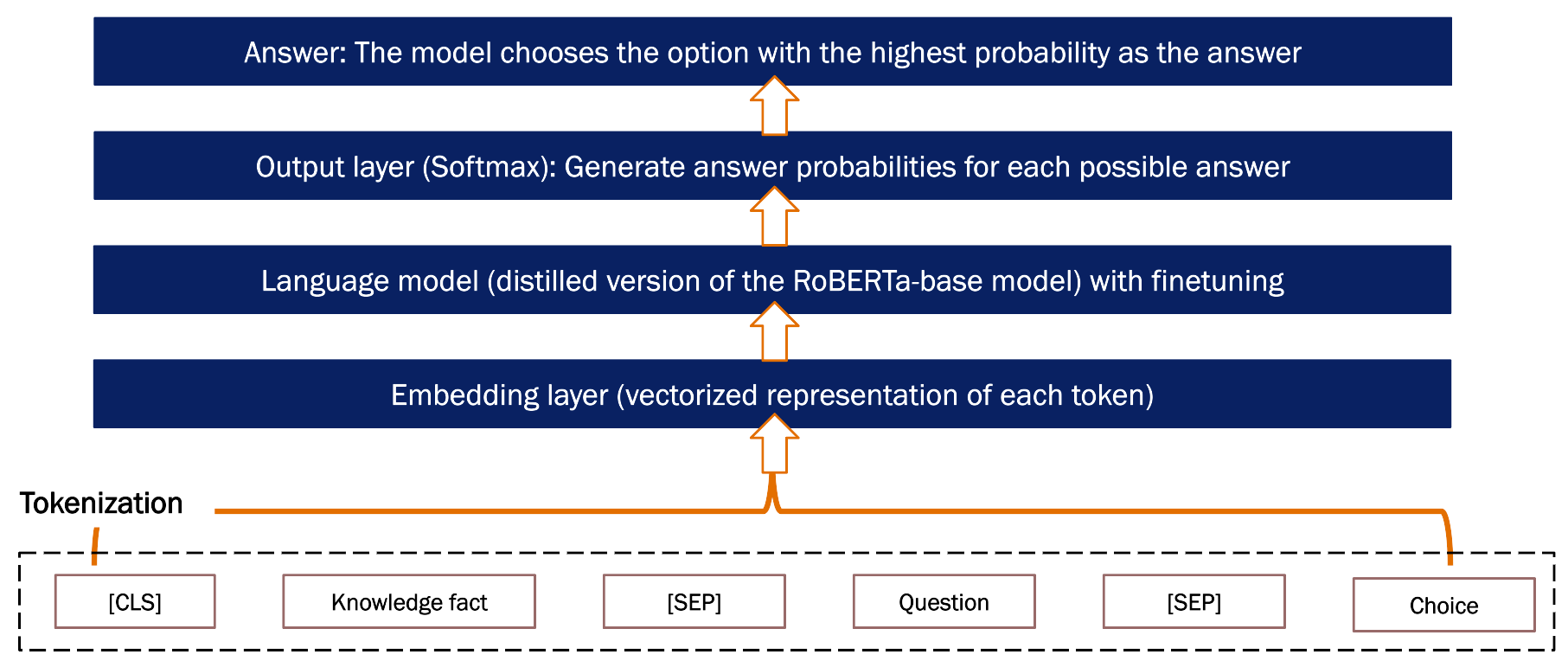}
  \caption{Approach 1: LM Architecture}
\end{figure}

\subsection{Approach 2: QAGNN method}
As described in Figure 2, the LM + KG method that we use is QAGNN \cite{qa-gnn}. QAGNN attempts to bridge the gap between the language and KG modalities by constructing a \textit{working graph} to embed the QA context and KG entities in the same representation space. It proposes to first embed each question and answer choice pair as a QA context vector, and also retrieve a subgraph from the ConceptNet KG to get k-hop neighbors of the entities in the question and answer choice. Then, it brings the two modalities in the same space by constructing a \textit{working graph} by treating the QA context vector as a node and relating it to each of the KG subgraph nodes with a relevance score, which is calculated using a pre-trained LM. This relevance score is then added as an additional feature for the node, and then an attention-based GNN module helps update all node representations for the \textit{working graph} in the same embedding space. Final predictions are then made using the LM representation, QA context node representation, and a pooled working graph representation. Note that we are not fine-tuning the LM module in the QAGNN method to stay consistent with \cite{qa-gnn}, in contrast to fine-tuning the LM model with the OpenbookQA dataset in approach 1.

Moreover, we test the impact of including supplementary knowledge facts in both the above-mentioned methods. Facts are incorporated in the QAGNN method by providing additional context in the subgraph extraction and relevance scoring steps. In the LM-only method, knowledge facts are concatenated with the QA context and used as an input to the modeling process. 

\begin{figure}[h]
  \centering
  \includegraphics[width=\linewidth]{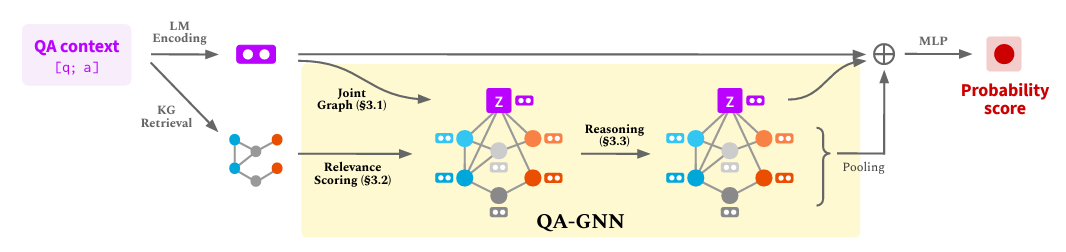}
  \caption{Approach 2: QAGNN Architecture}
\end{figure}

The final model was trained on Tesla V100 GPU (multi-GPU setting) and has the following hyperparameter setting:
\begin{itemize}
    \item Batch size:128
    \item Epochs: 20
    \item Encoder model: distilroberta-base
    \item Encoder learning rate: 1e-4
    \item GNN dimensions: 200
    \item Fully connected layer dimensions: 200
    \item GNN layers: 2/3/4
    \item Fully connected layers: 0/1
    \item Optimizer: Radam
    \item Loss: Cross-entropy
\end{itemize}

\section{Experiments and Results}


We use OpenbookQA dataset for training as per the split mentioned in Table 1. 
Due to computing constraints, we could not implement the models exactly as described in \cite{qa-gnn}. To overcome the computing constraints, we used distilled versions of the RoBERTa model and also used a lesser number of epochs and hops in the GNN. 
However, we still conducted experiments on the following hyperparameters:
\begin{itemize}
    \item GNN layers: We experiment with k= 2,3,4 to see the impact of increase in k-hop neighbours in GNN module.
    \item FC layers: We experiment with how the inclusion of a fully connected layer at output impacts the performance of the model.
    \item Knowledge fact context: We experiment with how including supplementary knowledge facts in addition to QA context impacts the performance of the model.
\end{itemize}
Note that we only experiment with adding knowledge fact context for the LM only approach, whereas we experiment with all the above-mentioned parameters for the QAGNN approach.

\subsection{Approach 1 results}

\begin{table}[!h]
\caption{LM model results (fine-tuned on OpenbookQA)} 
\begin{center}
\begin{tabular}{| c | c | c |}
\hline
\textbf{Model} & \multicolumn{2}{ c |}{\textbf{Accuracy}}  \\ 
\cline{2-3}
\textbf{setting} & \textbf{Dev} & \textbf{Test} \\
\hline

DistilRoBERTa-base(w/o facts) & 0.536 & 0.522  \\ \hline
DistilRoBERTa-base(with facts) & 0.658 & 0.670\\ \hline

\end{tabular}
\end{center}
\end{table}
The fine-tuned DistilRoBERTa-base model without knowledge fact context has an accuracy of 52.2\% whereas fine-tuned DistilRoBERTa-base model with ,knowledge fact context has accuracy of 67\%. Thus, we seean  significant improvement in performance on adding knowledge fact context. It aligns with our intuition that relevant context in addition to question-answer context, aids the model in making more informed predictions.

\subsection{Approach 2 results}

\begin{table}[!h]
\caption{QAGNN model results}

\begin{center}
\begin{tabular}{| c | c | c | c | c |}
\hline
\textbf{Model} & \textbf{GNN layers} & \textbf{FC} & \multicolumn{2}{ c |}{\textbf{Accuracy}}  \\ 
\cline{4-5}
\textbf{setting} & \textbf{(k)} & \textbf{layers} & \textbf{Dev} & \textbf{Test} \\
\hline

w/o facts & 2 & 0 & 0.526 & 0.484  \\ \hline
w/o facts & 2 & 1 & 0.534 & 0.498  \\ \hline
w/o facts & 3 & 0 & 0.552 & 0.516  \\ \hline
w/o facts & 3 & 1 & 0.546 & 0.486  \\ \hline

with facts & 2 & 0 & 0.650 & 0.648  \\ \hline
with facts & 2 & 1 & 0.636 & 0.640  \\ \hline
with facts & 3 & 0 & 0.642 & 0.662  \\ \hline
with facts & 3 & 1 & 0.652 & 0.676  \\ \hline
with facts & 4 & 0 & 0.652 & 0.682  \\ \hline
with facts & 4 & 1 & 0.669 & 0.694  \\ \hline

\end{tabular}
\end{center}
\end{table}
The QAGNN model without knowledge fact context, k=2, and no fully connected layer has test accuracy of 48.4\%. On observing the results in Table 4, we can make the following conclusions:
\begin{itemize}
    \item Model accuracy improves with an increase in GNN layers. QAGNN (w/o facts, k=2, FC layer = 0) has a test accuracy of 48.4\% as compared to QAGNN (w/o facts, k=3, FC layer = 0) has an accuracy of 51.6\%. Increasing the k-hop neighbors allows the GNN module to propagate messages from more distant nodes, thus incorporating topological information of the sub-graphs.
    \item Adding fulla y connected layer at output does not seem to improve the performance of the model. QAGNN ([w/o facts, k=2, FC layer = 0) has a test accuracy of 48.4\% as compared to QAGNN (w/o facts, k=2, FC layer = 1) has a test accuracy of 49.8\%, whereas QAGNN (w/o facts, k=3, FC layer = 0) has test accuracy of 51.6\% as compared to QAGNN (w/o facts, k=3, FC layer = 1) has test accuracy of 48.6\%. 
    \item Adding knowledge facts to the QA context drastically improves the performance of the model. For the most basic architecture QAGNN (w/o facts, k=2, FC layer = 0), the accuracy improves from 48.4\% to 64.8\% when knowledge fact context is included along with QA context. Adding context seems to make the module more informative and also seems to improve the relevance score of important nodes in the KG subgraph.
\end{itemize}

To gain a deeper understanding of the QAGNN method, we conducted a qualitative analysis comparing the model prediction with and without context. Our findings reveal that incorporating context significantly improves the model's predictions, as it enables a more comprehensive understanding of the underlying information. The addition of context aids in rectifying incorrect predictions by fostering a richer understanding of the question and answer space. Refer to Figures 3 and 4 for further details.

\begin{figure}[h]
  \centering
  \includegraphics[width=\linewidth]{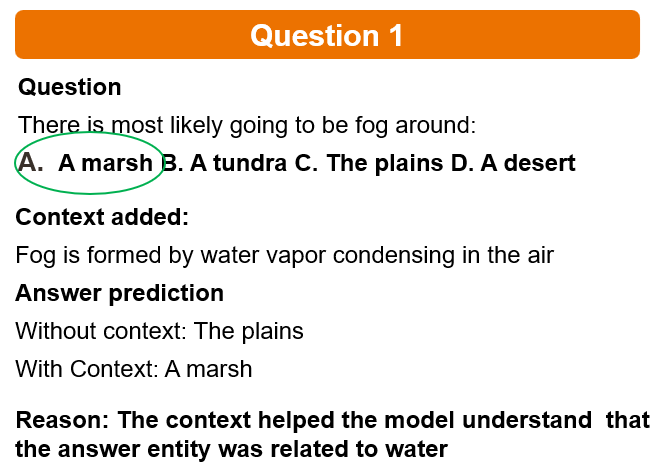}
  \caption{Sample question 1 deep-dive}
\end{figure}

\begin{figure}[h]
  \centering
  \includegraphics[width=\linewidth]{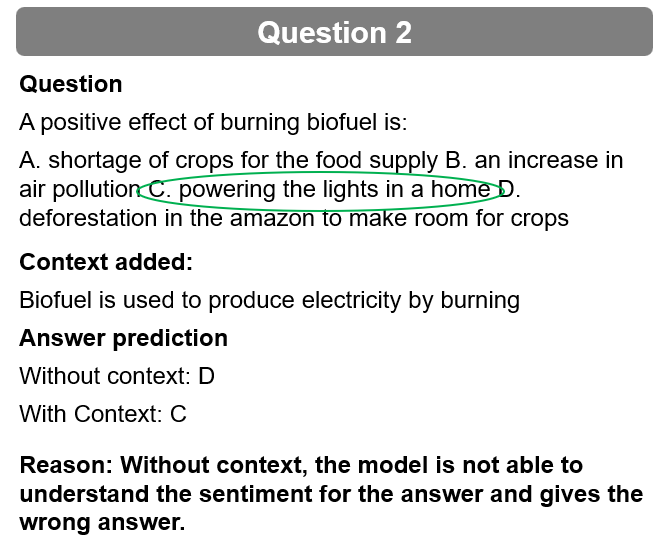}
  \caption{Sample question 2 deep-dive}
\end{figure}

\subsection{Comparative analysis between the methods}
For iterations with knowledge fact context, the LM model, with a test accuracy of 67\%, performs better than the simplest QAGNN architecture (with facts, k=2, FC layer =0) that has a test accuracy of 64.8\% but almost similar to another QAGNN architecture (with facts, k=3, FC layer =0) that has test accuracy of 66.2\%. Our best results are 69.4\% accuracy (QAGNN with facts, k= 4, FC layer =1).
Compared to the above results, \cite{qa-gnn} achieved the test accuracy of 82.8\% when using knowledge fact context. Although this result is significantly better than the results achieved by our group, the original paper has the following differences:
\begin{itemize}
    \item Base Model: The original work uses "AristoRoBERTa" for the modelling with knowledge facts as compared to "distilroberta-base". AristoRoBERTa is expected to perform better as it was trained on the OpenbookQA dataset. On the other hand, our group selected distilroberta-base as it has fewer parameters, almost similar performance as RoBERTa-base, and has much faster train times.
    \item GNN Layers: The original work use 5 layer GNN module as compared to 2/3/4 layer GNN module used by our group. As we observed above, an increase in GNN layers improves the performance, and we expect the original paper to have better performance.
    \item Epochs: The original work trained their QAGNN model over 100 epochs, which took ~20 hours. We limited our training to 20 epochs because of the computation limitations of Google Colab and storage limitations over PACE-COC clusters.
\end{itemize}

\section{Conclusion}
\subsection{Shortcomings of our work}
Although our QAGNN model generates satisfactory results with a test accuracy of 64.8\% even with the simplest architecture (k=2, FC layer = 0, and inclusion of knowledge fact context), there are a few limitations of our work:
\begin{itemize}
    \item Limit on input token size: As distilroberta-base model has a input token limit of 512 tokens, the overall QAcontext with knowledge fact context has a token size limit of 512. Our training, dev, and test dataset token sizes were well within this limit. However, if the QA + fact context size were larger, the model would truncate the overall input QA + fact context if the total token size is greater than 512 tokens.
    \item Model optimization: Our experimentation covers a small number of iterations than we'd like. It was limited in extensive hyperparameter tuning of different QAGNN architecture modules such as batch size, encoder model, learning rate, GNN dimensions, optimizer, etc. Thus, our work does not produce the most optimal model.  
    
\end{itemize}

\subsection{Extension of our work}

There are two major scopes of future work that we have outlined below:
\begin{itemize}
    \item KGs can be leveraged to study the explainability and structured reasoning leading up to the correct answer. To further understand explainability and the importance of KGs, we can perturb the question structure to check how the model behavior changes at inferencing.
    \item As seen in Table 4, we observe that supplementing facts with the OBQA dataset helps improve the accuracy of our model. We further propose to fuse the dataset with additional context such as Wiktionary data \href{https://en.wiktionary.org/wiki/Wiktionary:Main_Page}{[Data Link]} inspired by \cite{fusing-context}. 
    \item Integrating context from various sources may result in redundant information being included in the additional context. Additionally, numerous BERT-based models are constrained by a maximum input token size of 512 tokens. To prevent surpassing this limit when merging multiple contexts and effectively extracting the most valuable information, it is crucial to investigate efficient methods for combining context from diverse sources to enhance the original QA pair.
\end{itemize}

\printbibliography

\end{document}